\documentclass[preprint]{elsarticle}

\usepackage{amssymb,amsmath,epsfig,tabularx,delarray,enumerate,graphicx,array,xcolor,subcaption,amsthm,subfloat}
\usepackage{algorithm}
\usepackage{algorithmicx}
\usepackage{algpseudocode}
\usepackage[hidelinks]{hyperref}

\journal{Multimedia Tools and Applications}
\begin{document}
\begin{center}
\rule{\textwidth}{0.4pt}
{\small
\textbf{Citation Notice:} This is the preprint version of the article published in 
\textit{Multimedia Tools and Applications} (Springer, 2024).

Please cite the published version as:

Z. Farzadpour and M. Azghani, 
``Adaptive thresholding pattern for fingerprint forgery detection,'' 
\textit{Multimedia Tools and Applications}, vol. 83, pp. 81665–81683, 2024. 
\href{https://doi.org/10.1007/s11042-024-18649-3}{DOI: 10.1007/s11042-024-18649-3}

}
\rule{\textwidth}{0.4pt}
\end{center}

\begin{frontmatter}

\title{Adaptive Thresholding Pattern for Fingerprint Forgery Detection}

\author[mymainaddress]{Zahra Farzadpour}


\author[mymainaddress]{Masoumeh Azghani\corref{mycorrespondingauthor}}
\cortext[mycorrespondingauthor]{Corresponding author}

\address[mymainaddress]{Laboratory of Wireless Communication and Signal Processing (WCSP), Faculty of Electrical Engineering, Sahand University of Technology, Iran.}

\begin{abstract}
		Fingerprint liveness detection systems have been affected by spoofing, which is a severe threat for fingerprint-based biometric systems. Therefore, it is crucial to develop some techniques to  distinguish the fake fingerprints from the real ones. As fingerprint liveness detection systems play an important role in many security systems, and are unique for every person, spoofing these systems can result in dangerous consequences so much so that fingerprint image forgery detection research topic has drawn many researchers attention to it, and \color{black}in this paper, we propose a fingerprint forgery detection algorithm based on a suggested adaptive thresholding pattern. 
The anisotropic diffusion of the input image  is passed through three levels of the wavelet transform. The coefficients of different layers are adaptively thresholded	and concatenated to produce the feature vector which is classified using the SVM classifier. 	
Another contribution of the paper is to investigate the effect of various distortions such as pixel missing, block missing, and noise contamination. Our suggested approach includes a novel method that exhibits improved resistance against a range of distortions caused by environmental phenomena or manipulations by malicious users. In quantitative comparisons, our proposed method outperforms its counterparts by approximately $8\%$ and $5\%$ in accuracy for missing pixel scenarios of $90\%$ and block missing scenarios of size $70\times70$, respectively. This highlights the novelty and effectiveness of our approach in addressing these challenges.\color{black}
\end{abstract}

\end{frontmatter}

\renewcommand\thesection{{\Roman{section}}}
\section{Introduction}
Biometric recognition systems are an important element for recognizing people from each other. 
Nowadays,  concerning the advance of technology and the increase of security demands, the fingerprints have become popular as a unique and available recognition tool. A significant concern in biometric systems is their sensitivity to different attacks. These attacks can be set in one of the  following eight groups:
\begin{itemize}
	\item 1. Sensor levels: A fingerprint image is acquired throughout a camera by placing the finger on a transparent prism that can be fake.
	\item 2. Bypassing sensor: The sensor is ignored, and the previously stored data is replaced.\color{black} 
	\item 3. Feature Extraction: The attacker swaps the extracted features by the fake ones through trojan horse.
	\item 4. Feature substitution: Attacker substitutes feature vectors with wrong vectors.
	\item 5. Adopter: The attacker distorts the adopter and forces it to output preselected match scores.
	\item 6. Template data: The stored templates are altered by the attacker.
	\item 7. Corrupting template data: Transmitted template data are disrupted and damaged.
	\item 8. Disregarding result: The attacker overrides the final results.
\end{itemize}

Many software-based and hardware-based methods have been proposed for recognizing fingerprint forgery detection. However, hardware-based schemes need extra sensors to detect features like blood pressure, skin distortion, and skin odor. Hence, they are not efficient in comparison with the software-based methods \cite{chugh2017fingerprint}. Therefore, we focus on  the software-based methods.


 In this paper, we propose  a fingerprint forgery detection method. The motivation of the paper is to devise an algorithm which is resistant against various distortions. We have observed that the Anisotropic Diffusion (AD) \cite{perona1990scale} of a  fingerprint contains some clues to distinguish the fake fingerprints from the genuine ones. Therefore, the AD of the input image is computed and fed into a 3 level wavelet packet. The wavelet coefficients of  different layers are adaptively thresholded through a suggested adaptive thresholding algorithm and the results are combined, \color{black}using the binary pattern to obtain the proposed Adaptive Thresholding Patterns (ATP). The ATPs of the wavelet coefficients as well as the AD of the input image and the approximation coefficients of the first and second layers are concatenated to provide the feature vectors. \color{black}  At the last stage, the Support Vector Machine (SVM) classifier is applied to the feature vectors to identify the fake fingerprints. 
 
The simulation results have confirmed that the proposed method is resistant against various distortions such as pixel missing, block missing, and noise contamination. Moreover, the proposed method has higher detection accuracy compared to the other benchmarks in different scenarios. The proposed method achieves the accuracy of $0.88$ in the case of random pixel missing, while this value is $0.8$ for the best benchmark. Also, in the presence of the block missing of size $70\times 70$, the accuracy of the proposed method is about $0.85$, which is at least $0.1$ higher than those of the benchmarks. In the case of noise contamination with $SNR=-30 dB$, the suggested algorithm attains the accuracy of $0.96$, that is at least $0.04$ more than the accuracies of the benchmarks. 
  In the case of other pixel missing rates and block missing rates as well as various levels of noise contamination, the proposed method outperforms the other schemes from the detection accuracy perspective.

 The contributions of the paper can be summarized as:
 \begin{itemize}
	\item 1. An adaptive thresholding algorithm is suggested to produce a binary pattern for the fingerprint image.
	\item 2. The combination of the AD, wavelet transform and the adaptive thresholding pattern is suggested to extract efficient features of the fingerprint images.
	\item 3. The robustness of various algorithms against diverse distortions such as pixel missing, block missing, and noise contamination has been evaluated.
	\end{itemize}

 The rest of the paper is organized as follows: In Section II, the previously \color{black} related works in fingerprint forgery detection are discussed briefly. \color{black} In Section III, anisotropic diffusion \color{black}is illustrated.  Section IV provides a schematic of the proposed method to detect fingerprint spoofing. \color{black}Section V presents the experimental results, and the paper concludes \color{black}in Section VI.

\section{RELATED WORKS}
The software-based methods can be categorized into dynamic and  static schemes. The static methods require one fingerprint image, while the dynamic techniques demand several images acquired over time \cite{dubey2016fingerprint}. 
\renewcommand\thesubsection{{\Roman{subsection}}}
\subsection{Dynamic Methods}
Dynamic methods are divided into two groups based on skin deformation and perspiration. Skin deformation-based methods rely on the features of the skin surface. In \cite{jia2007new}, two features determined \color{black}human skin elasticity extracted from images captured over finger deformation. The first feature was \color{black}the correlation coefficient of the fingerprint region and signal intensity. The standard deviation of the fingerprint region length in the x and y axes was \color{black} the other feature.  In \cite{antonelli2006new}, the users were \color{black} asked to move their fingers on the sensor surface  to make deliberate distortions. Their results indicated \color{black} that the created distortion was \color{black} much more in the real fingerprints  than the fake ones.  In \cite{zhang2007fake}, the thin-plate spline model (TPS) had been incorporated \color{black} to compute the correlation of an input image with a real sample. The perspiration based methods rely on the fact that a real fingerprint indicates different moisture pattern compared to a fake one. The authors in \cite{parthasaradhi2005time} used the point that the sweat levels satiate around the pores and remain unchanged in a real fingerprint. In \cite{abhyankar2009integrating}, the changes over ridges of the perspiration patterns  had been investigated \color{black} using the wavelet transform to recognize the fake fingerprints. In \cite{nikam2009ridgelet}, the ridgelet energy and co-occurrence signatures had been used \color{black}to classify the fingerprint  textures. The dynamic methods require user cooperation and strongly  depend on the environmental factors such as temperature,  pressure, and dust on the sensor surface.
\subsection{Static Methods}
The static methods are divided into image quality based schemes and the perspiration pore based methods.  The image quality approaches use the statical features of the image texture.  In \cite{kulkarni2016fingerprint}, the Local Binary Pattern (LBP) with Discrete Shearlet transform had been used. \color{black}
 In \cite{ghiani2012fingerprint}, the Local Phase Quantization (LPQ) technique as well as the spectral characteristics had been utilized \color{black} to detect the fake fingerprints.
  In \cite{kim2015face},  the Local Speed Pattern (LSP) had been suggested \color{black} for face detection which  can also be used for fingerprint forgery detection.
%
     In \cite{kim2016fingerprint},  the Local Coherence Pattern (LCP) was applied \color{black} where
  the gradient intensity distribution had been used \color{black} to measure the directional coherence and dominant orientation.
%
 In \cite{chaudhari2012prevention}, the features like energy, entropy, median, variance, skewness, and kurtosis had been applied \color{black} to  detect the fake fingerprints.
 
The  Binarised Statistical Image Features (BSIF) were exploited \color{black}in \cite{ghiani2016fingerprint} as a local descriptor for fingerprint identification. 
  The discrete wavelet coefficients with Haar wavelet transform had been used \color{black}to calculate a binary pattern to extract features in \cite{shaju2017haar}. The Histogram of Oriented Gradient (HOG) was applied \color{black} in \cite{yuan2018fingerprint}. The gradient orientation of the real fingerprint is in one direction, while  it is in all directions in fake ones. In \cite{zaghetto2017liveness},  the Gray-level Co-occurrence Matrix (GLCM) was calculated \color{black}as the feature classified  with an artificial neural network. 
 In \cite{perona1990scale},  a method called Anisotropic Diffusion (AD)  was used \color{black}based on heat equation and Gaussian kernel function.  Also,  in another work \cite{liu2020fingerprint}, the combination of  AD and Shock Filtering (SF) was used \color{black}for sharpening the ridges of the fingerprints. Another technique used \color{black}Local Binary Pattern (LBP) and Harlick Transform  \cite{sabeena2021digital} to extract the features in digital images. The Uniform Local Binary Pattern (ULBP) with broad learning  \cite{chen2022broad} is a texture descriptor used in computer vision and image processing that encodes the relationship between the central pixel and its surrounding pixels in a binary code and counts the number of transitions from 0 to 1 or 1 to 0. The resulting code is then converted to decimal and transformed into a histogram for further analysis. In \cite{sharma2022intelligent}, Sharma and Selwal proposed an  Intelligent Fingerprint Presentation Attack Detection (IFPAD) scheme that extracts the local features of the image  based on the Local Adaptive Binary Pattern (LABP) and Uniform Local Binary patterns (ULBP). In another work \cite{agrawal2019fake}, the  local Haralick micro texture features together with macro features obtained from neighborhood gray-tone difference matrix (NGTDM) were exploited \color{black} to produce efficient feature vectors. In \cite{ogunlana2023impact}, Ogunlana presented a technique, involves image enhancement, singular point extraction, and post-processing stages, for detecting and eliminating noisy or false singular points in fingerprint images, improving the performance of fingerprint matching. Feng and Kumar in \cite{feng2023detecting} proposed a technique to extract fingerprint minutiae that uses a combination of a lightweight pixel-wise local neural network to extract local features and a deep patch-wise neural network, recovering global features. TruFor was suggested in \cite{guillaro2023trufor} as a forensic framework that detects and localizes image manipulations by extracting high-level and low-level traces and using a transformer-based fusion architecture. A novel Replay Attack Detection (RAD) scheme was presented in \cite{hamadouche2023replay} that combines perceptual image hashing and data hiding techniques to protect biometric systems against digital replay attacks, using a two-step verification process, including image authentication and identification, to detect possible fraudulent resubmission of biometric images. In another work \cite{baskar2023region}, a region-centric minutiae propagation measure (RCMPM) based forgery detection approach for fingerprint analysis in healthcare systems was implied. The technique involves preprocessing, extracting minutiae features, estimating RCMPM values based on neighboring regions, and classifying the fingerprints as genuine or forged. The technique proposed in \cite{hameed2023real}, uses texture features and classifiers to classify fingerprint images as authentic or altered. They used Histogram of Oriented Gradient (HOG) and Segmentation-based Feature Texture Analysis (SFTA) as input feature vectors.
 
 \color{black}It should be emphasized that all the fingerprint forgery detection schemes reviewed in this paper fall in the group of passive schemes where we have no access to the original image before its manipulation and the aim is to detect the forgery blindly. However, in the active forgery detection schemes such as the watermarking methods, a key is embedded inside the original image which enables the detection of the possible manipulation. The active forgery detection schemes are beyond the scope of this paper.  
  
\section{The Anisotropic Diffusion}
%

The Anisotropic Diffusion (AD) is illustrated in this section. The AD technique was originally suggested \color{black}for noise mitigation in image processing without omitting  the explanatory image parts such as edges and lines, which are notable for  image analysis \cite{perona1990scale}. 
In  AD, the aim is to have an approximation of the edges.  
  The AD of an input image $I_t$ is $I$ which was obtained \color{black}as \ref{equ2}: 



\begin{equation}
I_{i,j}^{t+1} = I_{i,j}^{t} + [c_{N_{i,j}}^{t}\cdot \bigtriangledown_{N}I_{i,j}^t + c_{S_{i,j}}^{t}\cdot \bigtriangledown_{S}I_{i,j}^t +
c_{E_{i,j}}^{t}\cdot \bigtriangledown_{E}I_{i,j}^t + c_{W_{i,j}}^{t}\cdot \bigtriangledown_{W}I_{i,j}^t]
\label{equ2}
\end{equation}
where the initial value of $I_{i,j}^{t}$ is $I_{i,j}^{0}=I_{i,j}$. In addition,  $N$, $S$, $E$, and $W$ represent the north, south, east, and west neighbourhood of the pixel,  $t$ is the number of iteration, and $\bigtriangledown_{N}, \bigtriangledown_{S}, \bigtriangledown_{E}, \bigtriangledown_{W}$ indicate the gradient in the north, south, east and west directions, respectively, which are defined as: 
\begin{equation}
\begin{aligned}
\bigtriangledown_{N}I_{i,j}^t  \equiv I_{i-1,j}^t - I_{i,j}^t \\
\bigtriangledown_{S}I_{i,j}^t  \equiv I_{i+1,j}^t - I_{i,j}^t \\
\bigtriangledown_{E}I_{i,j}^t  \equiv I_{i,j+1}^t - I_{i,j}^t \\
\bigtriangledown_{W}I_{i,j}^t  \equiv I_{i,j-1}^t - I_{i,j}^t
\label{equ3}
\end{aligned}
\end{equation}
The conduction coefficients $c_{N_{i,j}}^{t}, c_{S_{i,j}}^{t}, c_{E_{i,j}}^{t}, c_{W_{i,j}}^{t}$  are defined as an exponential function of the corresponding gradient:
\begin{equation}
\begin{aligned}
c_{N_{i,j}}^{t} = \exp (-(\frac{\bigtriangledown_{N}I_{i,j}^{t}}{\sigma})^{2}) \\
c_{S_{i,j}}^{t} = \exp (-(\frac{\bigtriangledown_{S}I_{i,j}^{t}}{\sigma})^{2}) \\
c_{E_{i,j}}^{t} = \exp (-(\frac{\bigtriangledown_{E}I_{i,j}^{t}}{\sigma})^{2}) \\
c_{W_{i,j}}^{t} = \exp (-(\frac{\bigtriangledown_{W}I_{i,j}^{t}}{\sigma})^{2}) 
\end{aligned}
\label{equ4}
\end{equation}
which $\sigma$ is the diffusion rate  considered as $\sigma=40$ in this paper.  In order to sustain the edges and sharpen the other regions, the conduction coefficients are chosen as an exponential function of the gradient of the brightness function. In this case, when the gradient is very high in region boundaries, the conduction coefficient becomes close to $0$ , and the AD equals the gradient of the image. If the gradient value is small, the AD of that region would be an updated average on the gradient due to the non-zero values of the conduction coefficients \cite{perona1990scale}. As a result, averaging is not applied to edges or corners; it only smoothens non-sharp regions. Therefore,  both sides of the image's edges would be smoothed while preserving edges that contain fine details. \color{black}

\begin{figure}
	\centering
	\includegraphics[scale=0.5]{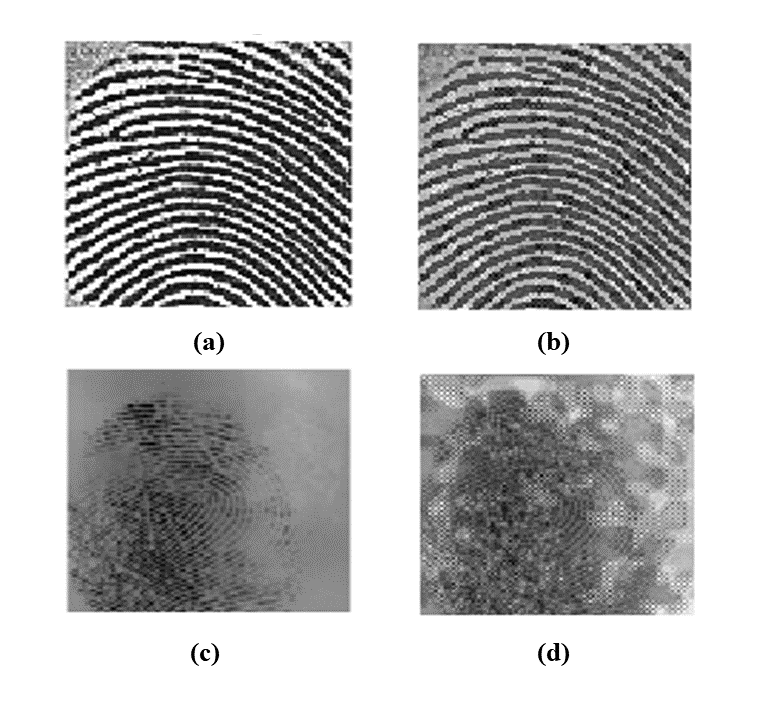}
	\caption{Anisotropic diffused images: (a) Real input image, (b) Real AD image, (c) Fake input image and (d) Fake AD image.}
	\label{fig:fig1}
\end{figure}

A fingerprint image and its AD have been presented in the \color{black}Figure \ref{fig:fig1}. It is obvious that the AD of a fake fingerprint image is visually different from that of a real one. Therefore, we would extract the features from the AD of a fingerprint.

\section{The Proposed Fingerprint Forgery Detection Scheme } 

\begin{figure}
	\centering
	\includegraphics[scale=0.5]{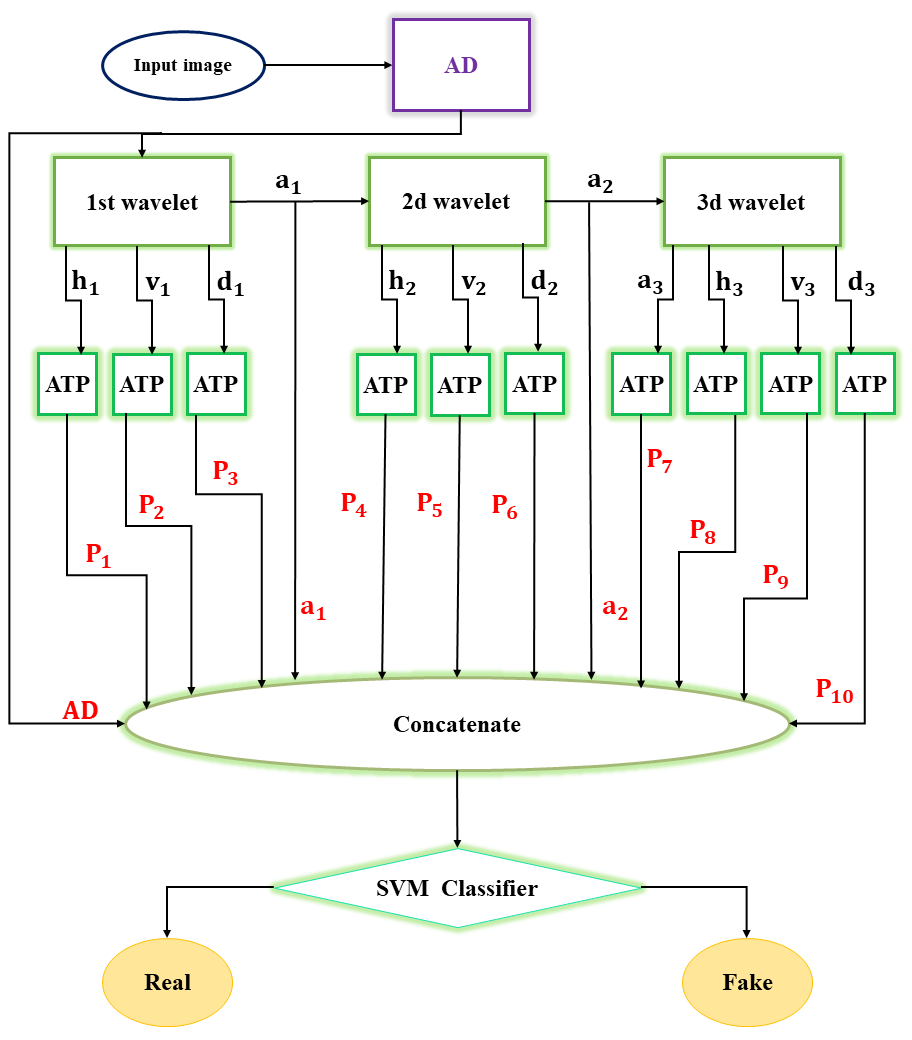}
	\caption{The flowchart of the proposed feature extraction technique.}
	\label{fig:fig3}
\end{figure}

The flowchart of the proposed fingerprint \color{black}forgery detection technique is represented in Figure \ref{fig:fig3}.

\begin{figure}
	\centering
	\includegraphics[scale=0.8
]{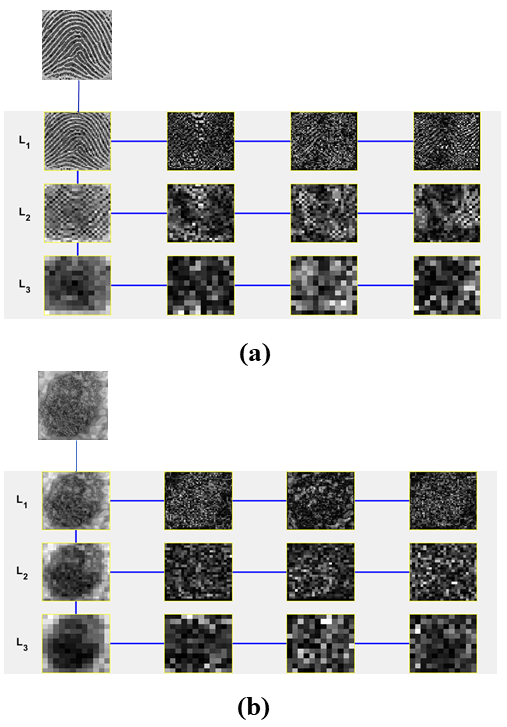}
	\caption{Three wavelet levels of the anisotropic diffused images: (a) The real image, (b) The fake image.}
	\label{fig:fig4}	 
\end{figure}

 The anisotropic diffusion is applied to the image. This is done since we aim to smoothen the image at the same time of preserving the edges and fine details  \cite{singh2021novel}. As the image is smoothened,  the factitious information appears after some repetitions. Then, the wavelet tranform is applied to the AD of the image in 3 levels (3 level wavelet packet), the subbands of all wavelet levels are shown in Figure \ref{fig:fig4}. We exploit the  Haar wavelet transform to localize the specific hidden points in the image. The Haar wavelet transform extracts the image's edge information, \color{black}which makes it a suitable choice for our proposed feature extraction scheme  \cite{olisa2018edge}, and the adaptive thresholding pattern (ATP) technique is used as LBP in \cite{thirumaleshwari2021multiple} for identifying micro-pattern compositions. \color{black}The first feature vector is the anisotropic diffusion of the input image, \color{black}represented as $\mathbf{AD}$. The approximation layer of the first and second levels are also used as two feature vectors ($\mathbf{a_1}$, $\mathbf{a_2}$). The approximation layer of each level is fed into a wavelet transform block to produce the four layers (approximation ($\mathbf{a_i}$), horizontal ($\mathbf{h_i}$), vertical ($\mathbf{v_i}$), and diagonal ($\mathbf{d_i}$), i=1,2,3) of the next level. The horizontal, vertical, and diagonal layers of the first and second level are fed into the ATP block to produce the other two feature vectors ($\mathbf{P_1}, \mathbf{P_2}, \mathbf{P_3}, \mathbf{P_4}, \mathbf{P_5}, \mathbf{P_6}$).   All of the four layers (the details as well as the approximation) of the third level are fed into the ATP blocks to provide the last feature vectors ($\mathbf{P_7}$, $\mathbf{P_8}$, $\mathbf{P_9}$, $\mathbf{P_{10}}$). The feature vectors ($\mathbf{AD},\mathbf{a_1},\mathbf{a_2},\mathbf{P_1}, \mathbf{P_2}, \mathbf{P_3}, \mathbf{P_4}, \mathbf{P_5}, \mathbf{P_6}, \mathbf{P_7}, \mathbf{P_8}, \mathbf{P_9}, \mathbf{P_{10}}$)  are concatenated to produce the total feature vector which is classified using an SVM classifier to determine whether the input image is a valid fingerprint \color{black}or a fake one. 

The ATP of a pixel considering its  $3\times 3$ neighborhood is derived as in Algorithm \ref{algorithm:algorithm1}.  
\begin{algorithm}
\caption {{The adaptive thresholding pattern (ATP)}}\label{algorithm:algorithm1} 
\begin{algorithmic}[1]
	\State \textbf{input:$\mathbf{I}\in\mathbb{R}^{M\times{N}}$}. 
	\State {The Number of thresholds:} $ K $.
	\State {The threshold values:} $T^{k}, k = 1 : K $
	\State {The neighborhood dimension:} $ m \times n $.
	\State\textbf{output:} $ \mathbf{R}\in\mathbb{R}^{M\times{N}} $
	    \For {$ m = 1 : M $}
	           \For {$ n = 1 : N $}
	                  \For {$ k = 1 : K $}
	                  \State $ ind = 0 $
	                                           \For {$ i = m-1 : m+1 $} 
	                             \For {$ j = n-1 : n+1 $}
	                               \State {$ ind \gets ind + 1 $}
	                                \State $ b^{k}(ind) \gets (\mathbf{I}(i,j) > T^{k})   $
	                             \EndFor
	                           \EndFor
	                           \State $ r^{k} \gets \sum_{i=1}^{8} b^{k}(i) 2^{i} $
	                 \EndFor
	               \State  $\mathbf{R}(m,n)\gets\sum_{k=1}^{K}r^{k}$
	            \EndFor
	    \EndFor
\end{algorithmic} 
\end{algorithm}

In the ATP algorithm, we consider $K$ threshold values, $T^{k}$. 
The $3\times 3$ neighbors of a pixel at location $(m,n)$  are compared with the $k^{th}$ threshold value $T^{k}$ to produce the entries of a $8$ element binary vector $b^k(.)$. A decimal value is calculated from the binary vector of each threshold, $r^k$. The sum of the $K$  decimal values produces the ATP of the corresponding \color{black}pixel $\mathbf{R}(m,n)$. 
The larger value of $r^k$ indicates that the neghiborhood of the pixel $(m,n)$ is greater than the $k^{th}$ threshold value, $T^{k}$. Using various thresholds in the ATP,  enables the algorithm to extract more information from the local areas by the comparison with diverse thresholding levels.
It is important to set proper threshold values in the ATP algorithm. For this purpose, we have suggested to determine the threshold values as follows:

  We choose a real sample image as a reference for tuning the thresholds. After applying AD on the reference image, its three-layer haar wavelet subbands are obtained. Then, for each pixel of the subbands,  the $k^{th}$ threshold value is computed as:
 \begin{equation}
 T^{k} (x,y) = \beta T^{0} (x,y) \exp (- \alpha (x,y) (k-1))
 \label{equ5}
 \end{equation}
 where $\beta$ is a constant value for all subbands. $T^{0}$ and $\alpha$ are calculated based on the $3 \times 3$ neighbourhood of each pixel as:
 \begin{equation}
 T^{0} (x,y) = H (x,y) - L (x,y)
 \end{equation}
 
 \begin{equation}
 \alpha (x,y) = \dfrac{H(x,y)}{T^{0}(x,y)}
 \end{equation}
where  $H(x,y)$ and $L(x,y)$ represent the highest and lowest values among the $3 \times 3$ neighbourhood of the  pixel $(x,y)$, respectively. In this manner, \color{black}various thresholds are generated based on the reference image for each subband. The obtained threshold values are used to evaluate all the fingerprints of the dataset. It should be noted that the selection of the reference image does not have a significant \color{black}change in the detection process. The advantages and disadvantages of our proposed method is indicated in Table \ref{tab:tab6}.\color{black}
\begin{table*}
	\caption{Pros and Cons of our proposed method.}
	\renewcommand{\arraystretch}{2}
	\centering  
	\begin{tabular}{| p{0.5\textwidth} | p{0.5\textwidth} | } 
		\hline
		\centering 
		\textbf{Pros} & \textbf{Cons}  \\
		\hline
		\hline
		\centering AD smoothes the image while preserving edges and details. & Robustness against unknown distortions is not guaranteed.  \\
		
		\centering Haar wavelet transform helps locate hidden points. &  \\
		
		\centering ATP can extract more information using diverse thresholds. & \\
		
		\centering Robust against different distortions. & \\
		\hline
	\end{tabular}
	\label{tab:tab6}
\end{table*}
 
%

\section{Experimental Result}
\begin{figure} 
	\centering
	\includegraphics[scale=0.7]{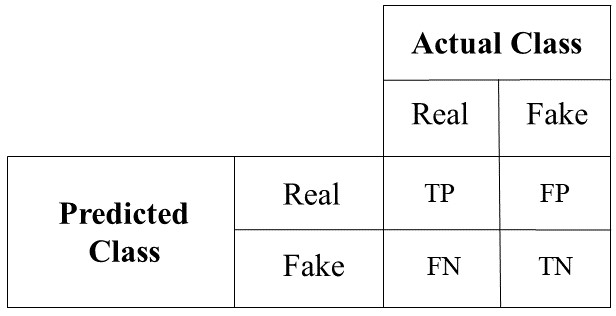}
	\caption{Confusion matrix diagram.}
	\label{fig:fig11}
\end{figure}

In this section, the simulation results are reported. The experiments are performed using MATLAB software on a  core i7 computer  with 6 GB RAM and a 2.4 GH processor. 
  The benchmarks of the proposed method are as follows: Anisotropic Diffusion (AD)  \cite{perona1990scale}, Histogram of Oriented Gradient (HOG)  \cite{yuan2018fingerprint}, Gray Level Co-occurrence Matrix (GLCM) \cite{zaghetto2017liveness}, Local Binary Pattern (LBP)  \cite{kulkarni2016fingerprint}, Local Coherence Pattern (LCP)  \cite{kim2016fingerprint}, Local Phase Quantization (LPQ)  \cite{ghiani2012fingerprint}, Local Binary Pattern and Harlick Transform (LBP\&HT) \cite{sabeena2021digital}, Uniform Local Binary Pattern (ULBP) \cite{chen2022broad}, Local Adaptive Binary Pattern and Uniform Local Binary patterns (LABP\&ULBP) \cite{sharma2022intelligent}. 
The comparison measures are \color{black}Accuracy, Precision, Recall, and F1-Score. The  Accuracy is defined as follows:

\begin{equation}
Accuracy = \dfrac{(TP+TN)}{TP+TN+FP+FN}
\end{equation}
where TP, TN, FP, and FN are true positive, true negative, false positive, and false negative detections, respectively. These values are shown in a confusion matrix in Figure \ref{fig:fig11}.
\color{black}Precision means the percentage of the truly \color{black}estimated results with respect to the whole of the positive ones:
\begin{equation}
Precision = \dfrac{TP}{TP+FP}
\end{equation}
On the other hand, Recall refers to the ratio of the retrieved and relevant instances to all of the relevant \color{black}ones:
\begin{equation}
Recall= \dfrac{TP}{TP+FN}
\end{equation}
F1-score is a more straightforward metric to evaluate the system's \color{black}performance regarding precision and recall. This metric is simply the harmonic mean of precision and recall which evaluates the effect of both simultaneously.
 \begin{equation}
F1-score = 2 \times\dfrac{Precision \times Recall}{Precision + Recall}
\end{equation}

For the ATP block, we have applied five different thresholds for each subband ($K=5$). We also set the parameter $\beta$ in \eqref{equ5} to $2.2$ to achieve the best performance of our proposed method. The obtained thresholds are given in Table \ref{tab:tab1}. It should be noted that these values have been adjusted by trial and error on different layers of various image samples. The important intuition behind the selection of the thresholds is that the values are decreasing iteration by iteration.
\begin{table*}
	\caption{The threshold values.}
	\centering  
	\renewcommand{\arraystretch}{2}
	\begin{tabular} {| p{0.3\textwidth} | p{0.15\textwidth} | p{0.15\textwidth} | p{0.15\textwidth} |p{0.15\textwidth} |p{0.15\textwidth} |} 
		\hline
		\centering 
		\textbf{Subband} & \textbf{$T^{0}$} & \textbf{$T^{1}$} & \textbf{$T^{2}$} & \textbf{$T^{3}$} & \textbf{$T^{4}$} \\
		\hline
		\hline
		\centering $h_{1}$ & 547.5502 & 350.2337 & 224.0222 & 143.2933 & 91.6558 \\
		\hline
		\centering $v_{1}$ & 242.1100 & 133.3524 & 73.4495 & 40.4554 & 22.2825 \\
		\hline
		\centering $d_{1}$ & 290.7339 & 150.5746 & 77.9844 & 40.3891 & 20.9180 \\
		\hline
		\centering $h_{2}$ & 1064.9976 & 688.9769 & 445.7185 & 288.3478 & 186.5403 \\
		\hline
		\centering $v_{2}$ & 411.0094 & 262.3547 & 167.4658 & 106.8964 & 68.2339 \\
		\hline
		\centering $d_{2}$ & 257.9083 & 162.0467 & 101.8158 & 63.9720 & 40.1944 \\
		\hline
		\centering $a_{3}$ & 904.0261 & 53.7971 & 3.2014 & 0.1905 & 0.0113 \\
		\hline
		\centering $h_{3}$ & 699.1689 & 430.6025 & 265.1985 & 163.3298 & 100.5912 \\
		\hline
		\centering $v_{3}$ & 330.3916 & 178.6908 & 96.6441 & 52.2696 & 28.2698 \\
		\hline
		\centering $d_{3}$ & 594.4882 & 420.6886 & 297.6996 & 210.6666 & 149.0779 \\
		\hline
	\end{tabular}
	\label{tab:tab1}
\end{table*}

We used the ATVS-FFp DB dataset \cite{galbally2012high, galbally2011evaluation} which consists of  real and fake fingerprints. The ATVS-FFp DB is a collection of high-quality fingerprint images that are commonly used for fingerprint recognition research. It includes over 1,200 fingerprint images from 125 different individuals, with each individual contributing between 5 and 16 various fingerprints. The images are saved in \emph{.bmp} format with a resolution of 500 dpi and include both plain and rolled fingerprints.  We select 120 fingerprint images of different sensors for the training phase, 60 real samples and 60 fake samples. Also, for the testing phase, we choose 400 samples of various sensors, 300 real and 100 fake samples. 

All the benchmarks as well as the proposed method have been applied on the dataset to detect the forged fingerprints.
The Accuracy, Precision, Recall, and F1-score measures of all the methods are compared in Table \ref{tab:tab2}.

\begin{table*}
	\caption{The fingerprint test results on the ATVS-FFp DB dataset. }
	\centering  
	\renewcommand{\arraystretch}{2}
	\begin{tabular}{| p{0.4\textwidth} | p{0.15\textwidth} | p{0.15\textwidth} | p{0.15\textwidth} |p{0.15\textwidth} |p{0.15\textwidth} |} 
		\hline
		\centering 
		\textbf{Method} & \textbf{Accuracy} & \textbf{Precision} & \textbf{Recall} & \textbf{F1-score} \\
		\hline
		\hline
		\centering Proposed method & 1.00 & 1.00 & 1.00 & 1.00 \\
		\hline
		\centering AD & 1.00 & 1.00 & 1.00 & 1.00 \\
		\hline
		\centering LBP & 1.00 & 1.00 & 1.00 & 1.00 \\
		\hline
		\centering LPQ & 1.00 & 1.00 & 1.00 & 1.00 \\
		\hline
		\centering GLCM & 0.9925 & 0.9899 & 0.9800 & 0.9849 \\
		\hline
        \centering HOG & 0.9875 & 0.9897 & 0.9600 & 0.9746 \\
        \hline
        \centering LABP\&ULBP & 0.9325 & 0.8120 & 0.95 & 0.8756  \\
        \hline
        \centering  ULBP & 0.9300 & 0.9865 & 0.73 & 0.8391 \\
        \hline
        \centering  LBP\&HT & 0.8575 & 0.7129 & 0.72 & 0.7164 \\
        \hline
        \centering LCP & 0.8375 & 0.6259 & 0.8700 & 0.7280 \\
        \hline
	\end{tabular}
	\label{tab:tab2}
\end{table*}

According to the results, the proposed method, AD, LBP, and LPQ methods have the best F1-scores as well as the precision and recall values. The LCP scheme has the worst performance among all the methods.

Furthermore, we aim to investigate the robustness of the algorithms in various distortions. This is of paramount importance as various distortions might be applied to the forged fingerprint, misleading \color{black}the authentication systems. The three distortions considered in this paper are as follows: random pixel missing, block missing, and the  additive white gaussian noise (AWGN). 

\begin{figure} [H]
	\centering
	\includegraphics[scale=1]{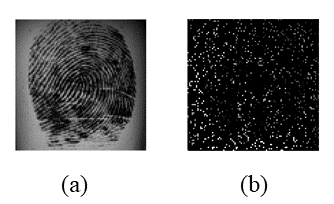}
	\caption{A sample of fingerprint image and its distorted version  with  $90\%$ random pixel missing rate.}
	\label{fig:fig5}
\end{figure}

As the first distortion, the random pixel missing is considered. A random binary mask is generated which is inner producted by the input image to produce the random pixel missing image. Therefore, some of the pixels of the image would be zero (missing). This kind of distortion might happen during the transmission of the fingerprint image over a network due to packet loss, or the signal capture \color{black}due to the malfunctioning of some of the sensors, or it can even occur intentionally by the forgers for malicious purposes. A sample of random pixel missing distortion has been depicted in Figure \ref{fig:fig5}.
The forgery detection results for the case of $90\%$ pixel missing rate have been given in Table \ref{tab:tab3}. In this case, $90\%$ of the pixels have been missed according to a random pattern. The average of 4 MontCarlo runs has been given in this table.

 \begin{table*}
	\caption{The random pixel missing results for $90\%$  missing rate.}
	\renewcommand{\arraystretch}{2}
	\centering  
	\begin{tabular} {| p{0.3\textwidth} | p{0.15\textwidth} | p{0.15\textwidth} | p{0.15\textwidth} |p{0.15\textwidth} |p{0.15\textwidth} |}  
		\hline
		\centering 
		\textbf{Method} & \textbf{Accuracy} & \textbf{Precision} & \textbf{Recall} & \textbf{F1-score} \\
		\hline
		\hline
		\centering Proposed method & 0.884375 & 0.761500 & 0.787500 & 0.773025 \\
		\hline
		\centering AD & 0.801250 & 0.557150 & 1.00 & 0.715550\\
		\hline
		\centering GLCM, HOG, LBP,  ULBP,  LBP\&HT, LABP\&ULBP, LCP and LPQ & 0.7500 & - & 0.00 & - \\
		\hline
	\end{tabular}
	\label{tab:tab3}
\end{table*}
  The proposed method outperforms the AD algorithm in detecting the fake forgeries. Comparing the F1-scores, we observe that the proposed method acts almost 5.7 percent better than the AD scheme. Based on the precision and recall values, we can say that our technique \color{black}operates 4 percent better than the AD scheme. The rest of the benchmarks fail to detect \color{black}the fake fingerprints in the presence of random pixel missing distortion, as \color{black}there is no positive detection. Therefore, their Precision and F1-score are invalid values and the Recall  is zero.\color{black}

\begin{figure} [H]
 	\centering
 	\includegraphics[width=1\linewidth]{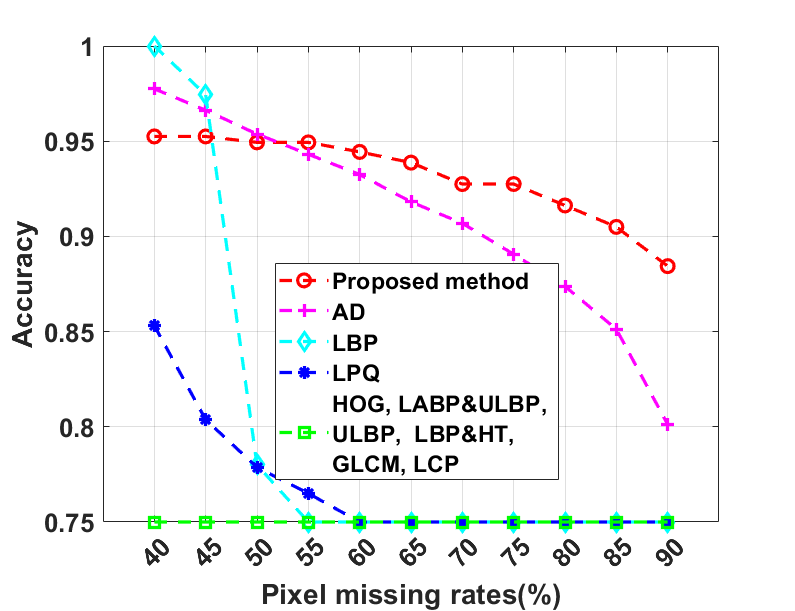}
 	\caption{Comparison of the Accuracy of all the methods versus different pixel missing rates }
 	\label{fig:fig6}
 \end{figure}

  In Figure \ref{fig:fig6}, we have compared all the methods in different pixel missing rates changing from $40\%$ to $90\%$. 
  In the minimum pixel missing rate of $40\%$, LBP has the maximum accuracy of $1$. At the same time, this value for our proposed method is about $0.95$, \color{black}which is $2.5\%$ lower than AD. Although LBP and AD methods have slightly better performance in the lower missing rates, our proposed method indicates more robustness against higher pixel missing rates. 
 Furthermore, it is clear that the HOG, LABP\&ULBP, ULBP, LBP\&HT, GLCM, and LCP methods cannot recognize the fake fingerprints in all pixel missing rates. Also, the LBP and LPQ schemes have the same behavior in forgery detection after some pixel missing rates. 
   
\begin{figure} [H]
	\centering
	\includegraphics[scale=1]{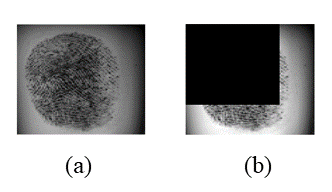}
	\caption{A sample of block missing distortion on a fingerprint image  by eliminating a $70\times 70$ block.}
	\label{fig:fig7}
\end{figure}

    In this part, the block missing distortion is applied to \color{black}the fingerprint images. To this end, a $70\times 70$ block of the input  data of size $96\times 96$ has been missed, replaced with a zero block of the same size. We defined a fixed location for the missing block in all of the test images. This kind of distortion might occur due to the wrong placement of the finger on the sensors, or a deliberate deceptive action.  An example of a block missing distortion has been presented in Figure \ref{fig:fig7}.
   
  The detection results of all the methods have been described in Table \ref{tab:tab4} for deleting a $70\times 70$ block of the test data.
  
\begin{table*}
	\caption{The block missing results by eliminating a $70\times 70$ block of the test data.}
	\renewcommand{\arraystretch}{2}
	\centering  
	\begin{tabular}{| p{0.3\textwidth} | p{0.15\textwidth} | p{0.15\textwidth} | p{0.15\textwidth} |p{0.15\textwidth} |p{0.15\textwidth} |} 
		\hline
		\centering 
		\textbf{Method} & \textbf{Accuracy} & \textbf{Precision} & \textbf{Recall} & \textbf{F1-score} \\
		\hline
		\hline
		\centering Proposed method & 0.8525 & 1.00 & 0.4100 & 0.5816 \\
		\hline
		\centering LBP & 0.7550 & 1.00 & 0.0200 & 0.0392 \\
		\hline
		\centering HOG, LABP\&ULBP, ULBP, LBP\&HT, and LPQ & 0.7500 & - & 0.00 & - \\
		\hline
		\centering LCP & 0.7300 & 0.4808 & 1.00 & 0.6494 \\
		\hline
		\centering GLCM & 0.7025 & 0.4566 & 1.00 & 0.6270 \\
		\hline
		\centering AD & 0.4725 & 0.3215 & 1.00 & 0.4866 \\
		\hline
	\end{tabular}
	\label{tab:tab4}
\end{table*}

In the block missing distortion, our method achieves at least  9\% higher Accuracy than the other schemes. Also, the precision of the proposed method is the perfect value of 1. However, in the case of the Recall value, the proposed method is inferior to \color{black}the LCP, GLCM, and AD  methods.  This is because our method failed to disclose \color{black}some of the fake samples.  The HOG, LABP\&ULBP, ULBP, LBP\&HT, and LPQ schemes fail in detecting the fake fingerprints, so their Precision and F1-score are not valid values.\color{black}
 Therefore, from the F1-score point of view, the proposed method outperforms all the methods except LCP and GLCM. This is so because the proposed method's Recall value is lower than those of LCP and GLCM.\color{black}
 
\begin{figure} [H]
	\centering
	\includegraphics[width=1\linewidth]{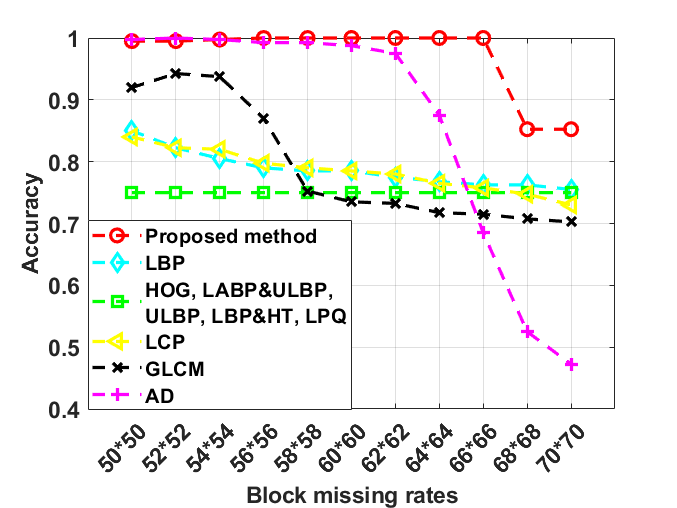}
	\caption{The comparison of various methods with different block missing rates.}
	\label{fig:fig8}
\end{figure}

We also examined techniques based on various block missing rates from $50\times50$ to $70\times70$ block sizes in Figure \ref{fig:fig8}. In the missing block of $50\times50$,  our proposed method as well as the AD scheme have the best performance compared to the others, \color{black}which have accuracies lower than $0.975$.

It is evident that the rival methods have the least resistance to the larger missing block sizes. Additionally, the AD scheme could not demonstrate \color{black}acceptable performance at the large missing block size and offered the lowest Accuracy value.  In brief, all the Accuracies decreased as the missing block size increased, \color{black}and our proposed method depicts the highest \color{black}reliability in this type of distortion.

\begin{figure}[H]
	\centering
	\includegraphics[scale=1]{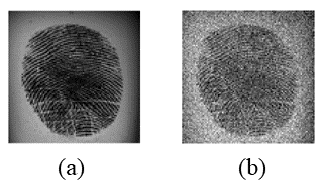}
	\caption{A sample fingerprint image contaminated with the AWGN noise of  $SNR=-30 dB$. }
	\label{fig:fig9}
\end{figure}

 In the last type of distortion,  the test images are contaminated  with additive white gaussian noise with an SNR of -30 dB. This kind of distortion might occur due to the sensing system's internal noise or a deceptive action by forgers to confuse the authentication system. A sample of the fingerprint image contaminated  with the AWGN noise at an SNR of -30 dB  is shown in Figure \ref{fig:fig9}. \color{black} 
The comparison measures of all the methods in the presence of this distortion have been given in Table \ref{tab:tab5}.
\begin{table*}
	\caption{The results of the AWGN distortion for  $SNR=-30 dB$.}
	\renewcommand{\arraystretch}{2}
	\centering  
	\begin{tabular}{| p{0.3\textwidth} | p{0.15\textwidth} | p{0.15\textwidth} | p{0.15\textwidth} |p{0.15\textwidth} |p{0.15\textwidth} |} 
		\hline
		\centering 
		\textbf{Method} & \textbf{Accuracy} & \textbf{Precision} & \textbf{Recall} & \textbf{F1-score} \\
		\hline
		\hline
		\centering Proposed method & 0.968750 & 0.926848 & 0.9500 & 0.938264 \\
		\hline
		\centering LPQ & 0.923750 & 1.00 & 0.6950 & 0.819875 \\
		\hline
		\centering AD & 0.920625 & 0.815225 & 0.882500 & 0.847575 \\
		\hline
		\centering LBP & 0.869375 & 0.656900 & 1.00 & 0.792900 \\
		\hline
		\centering GLCM, LABP\&ULBP,  ULBP, and LBP\&HT  & 0.7500 & - & 0.00 & - \\
		\hline
		\centering HOG & 0.6550 & 0.420175 & 1.00 & 0.591725 \\
		\hline
		\centering LCP & 0.64375 & 0.00 & 0.00 & - \\
		\hline
	\end{tabular}
	\label{tab:tab5}
\end{table*}

 In this table, the proposed method has almost  0.9700  Accuracy, while the accuracy of the best rival method, LPQ, is 0.92375 and the other methods have lower Accuracy.  \color{black}The GLCM, LABP\&ULBP, ULBP, and LBP\&HT schemes did not show any robustness in the case of AWGN noise and they fail in distinguishing the fake data from the real one. Based on Precision and Recall values in Table \ref{tab:tab5}, our proposed method is better in detecting real samples while the LPQ is better in identifying the fake ones.  In this case, LCP puts some of the real samples in the fake category and all the fake ones in the real category, so the precision and recall have zero values. Overall, from the F1-score point of view, the proposed method has better performance than its rivals in the case of noisy test images. Its F1-score is at least 10 percent higher than the other schemes. 
  
\begin{figure} [H]
 	\centering
 	\includegraphics[width=1\linewidth]{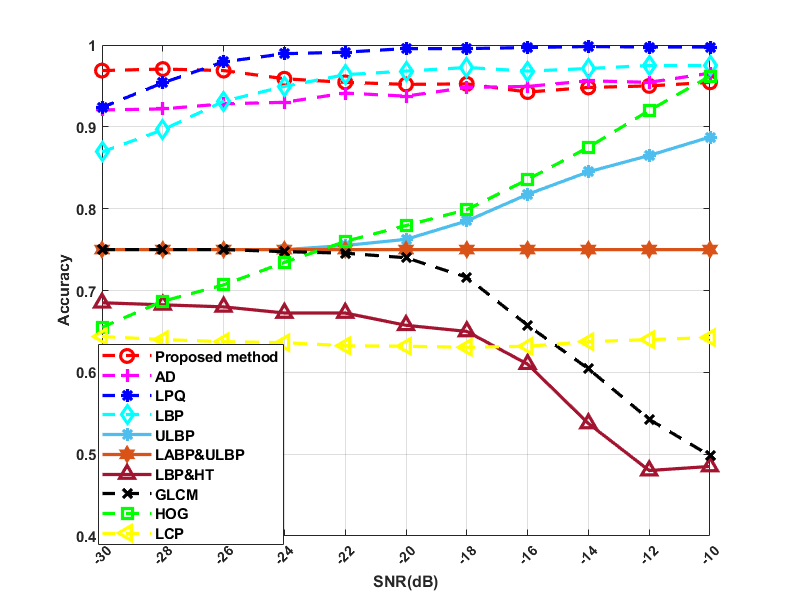}
 	\caption{The Comparison of the methods with AWGN distortion for different SNR values. }
 	\label{fig:fig10}
 \end{figure}

 In Figure \ref{fig:fig10}, the Accuracies of all the methods based on various SNR values ranging from $-10 dB$ to $-30 dB$ are illustrated, \color{black}where $SNR= -30 dB$ has the largest distortion.
  In general, the Accuracies of most methods increase with SNR. However, our method, AD, and LCP exhibit small oscillations due to the inherent randomness of AWGN distortion. In addition, the GLCM and LBP\&HT schemes show a different trend compared to other techniques, with their accuracy decreasing as SNR increases. 
 
 At lower SNR values, LPQ, LBP, and AD have the best accuracies chronologically, while after the SNR of $-26 dB$, our proposed method surpasses the rivals and reaches an Accuracy of about $0.97$.
 
 Regarding other methods, HOG's accuracy significantly improves from $0.655$ at an SNR of -30 dB to $0.96$ at an SNR of -10 dB. Meanwhile, LCP experiences a marginal change from $0.63$ to $0.64$.

In contrast, the GLCM and LBP\&HT methods exhibit contradictory changes compared to other methods. Their accuracies drop from approximately $0.75$ to $0.5$ and $0.6850$ to $0.4850$ respectively. This can be attributed to the nature of these methods which rely on pixel value similarity. The larger AWGN distortion may increase this similarity and mislead these techniques in accurate data classification.\color{black}

\section{Conclusion}
In this paper, we suggested an algorithm for  fingerprint forgery detection. The suggested algorithm relies on the intuition that the anisotropic diffusion of the fake fingerprint is different from that of the real one. Therefore, we develop a technique to extract the features of the AD image.  To this goal, three level wavelet transform is applied to the anisotropic diffusion of the fingerprint image. Then, the coefficients of different layers are fed into a proposed adaptive thresholding block to produce the features which are classified using the SVM scheme.  
The introduced algorithm has been compared to its counterparts in various scenarios where different distortions are applied to the fingerprints. The proposed method shows more robustness against pixel missing, block missing, and noise contamination distortions, in comparison with the rivals. As the future work, other feature extraction techniques can be developed to provide much higher robustness against various distortions. Furthermore, the deep learning based techniques can be exploited for the fingerprint forgery detection. 
\section{Funding and Conflicts of interests}
The authors declare that they have no known competing financial interests, personal relationships, and conflicts of interest that could have appeared to influence the work reported in this paper.

\bibliographystyle{IEEEtran}
\bibliography{cite}

\end{document}